%% file: main.tex
\documentclass[conference]{IEEEtran}
\IEEEoverridecommandlockouts
\usepackage{cite}
\usepackage{graphicx,epstopdf,url}
\usepackage{amsmath}
\usepackage{amssymb}
\usepackage{amsfonts}
\usepackage{graphicx}
\usepackage{verbatim}
\usepackage{textcomp}

\usepackage{booktabs}
\usepackage[utf8]{inputenc}

\usepackage{xcolor}
\usepackage{soul}
\usepackage{tabularx}
\usepackage{makecell}
\usepackage{wrapfig}
\usepackage{footnote}
\usepackage{bbm}
\usepackage{algorithm}
\usepackage{algpseudocode}
\usepackage{subcaption}
\usepackage{indentfirst}
\usepackage[all]{xy}
\usepackage{bbm}
\newcommand{\our}{DIMEE}

\def\BibTeX{{\rm B\kern-.05em{\sc i\kern-.025em b}\kern-.08em
    T\kern-.1667em\lower.7ex\hbox{E}\kern-.125emX}}
\begin{document}

\title{Distributed Inference on Mobile Edge and Cloud: An Early Exit based Clustering Approach}


\author{\IEEEauthorblockN{Divya Jyoti Bajpai and Manjesh Kumar Hanawal}
\IEEEauthorblockA{\textit{Dept. of IEOR, IIT Bombay} \\
Mumbai, Maharashtra-400076, India\\
\{divyajyoti.bajpai, mhanawal\}@iitb.ac.in}
}

\maketitle

\begin{abstract}
Recent advances in Deep Neural Networks (DNNs) have demonstrated outstanding performance across various domains. However, their large size is a challenge for deployment on resource-constrained devices such as mobile, edge, and IoT platforms. To overcome this, a distributed inference setup can be used where a small-sized DNN (initial few layers) can be deployed on mobile, a bigger version on the edge, and the full-fledged on the cloud. A sample that has low complexity (easy) could be then inferred on mobile, that has moderate complexity (medium) on edge, and higher complexity (hard) on the cloud. As the complexity of each sample is not known beforehand, the following question arises in distributed inference: how to decide complexity so that it is processed by enough layers of DNNs. We develop a novel approach named \our{} that utilizes Early Exit (EE) strategies developed to minimize inference latency in DNNs. \our{} aims to improve the accuracy, taking into account the offloading cost from mobile to edge/cloud. Experimental validation on GLUE datasets, encompassing various NLP tasks, shows that our method significantly reduces the inference cost $(>43\%)$ while maintaining a minimal drop in accuracy $(<0.3\%)$ compared to the case where all the inference is made in cloud. \footnote{The source code is available at \url{https://github.com/Div290/DIMEE}}
\end{abstract}

\begin{IEEEkeywords}
Early Exit, Distributed inference.
\end{IEEEkeywords}

\section{Introduction}

\input{chapters/intro}

\input{chapters/related_works}

\input{chapters/problem_setup}

\input{chapters/experiments}

\input{chapters/results}

\input{chapters/conclusion}

\bibliographystyle{IEEEtran}
\bibliography{mab}

\end{document}

%% file: chapters/intro.tex
In recent years, Deep Neural Networks (DNNs) have significantly increased in scale, resulting in outstanding performance \cite{han2022survey}, particularly in Natural Language Processing (NLP) \cite{wang2019glue} tasks. However, this growth in scale necessitates substantial computational resources, which restricts their deployment on resource-constrained platforms like mobile and edge devices. To address these challenges, various strategies have been proposed, including model pruning, weight quantization, knowledge distillation, early exits, and cloud offloading \cite{matsubara2022split}.

Methods such as model pruning \cite{zhu2017prune, michel2019sixteen} weight quantization \cite{zhang2020ternarybert, kim2021bert} and knowledge distillation \cite{sanh2019distilbert, jiao2019tinybert} tend to lower the model size by different methods, significantly reducing the accuracy of the models. These methods mostly compress the model that can fit in the memory of the mobile device but affect the optimality of the backbone. Some models also provide smaller versions of their large-sized models \cite{devlin2018bert, carreira2023revolutionizing} to fit them in resource-constrained devices.


As mobile and edge devices often lack the capability to perform inference on large models due to resource constraints such as limited space and memory, cloud offloading leverages high-capacity services and extensive computing resources, allowing the deployment of full-fledged DNNs for inference. However, offloading samples to the cloud incurs additional costs due to the physical distance from mobile terminals. Moreover, not all samples require the same amount of computation since real-world datasets comprise a mixture of easy and hard samples.

To address this, we utilize distributed inference: deploying initial layers of the DNN on the mobile device, a larger model with more layers on the edge device, and the full model on the cloud.
Given the varying complexity of real-world samples, it is advantageous to utilize mobile, edge, and cloud resources based on the complexity of incoming samples. As the complexity of the incoming samples is unknown, the question arises how to identify it. We address this challenge of identifying the complexity of the sample so that one can decide whether the sample is to be inferred at the mobile, edge, or cloud. Further, a sample inferred in the cloud can have better accuracy but involves higher offloading costs. On the other hand,  inferring all the samples on mobile can degrade accuracy. Hence, the decision of where to infer has to account for both accuracy and offloading cost. 

Recently, Early Exit (EE) strategies have gained attention for adaptive inference \cite{teerapittayanon2016branchynet, xin2020deebert}, where inference can be made at classifiers attached at the intermediary layer. The primary goal of EE strategies is to reduce inference latency by letting the sample exit from the intermediary layer if the prediction confidence at that layer exceeds a predefined threshold. The initial layers of DNNs extract low-level features sufficient for easy samples, while deeper layers are required for more complex features necessary for harder samples. 
Allowing easier samples to exit early reduces computational demand and increases inference speed. 
EE strategies perform inference based on sample complexity, making them ideal for distributed inference scenarios.



\begin{figure*}
    \centering
    \includegraphics[scale=0.59]{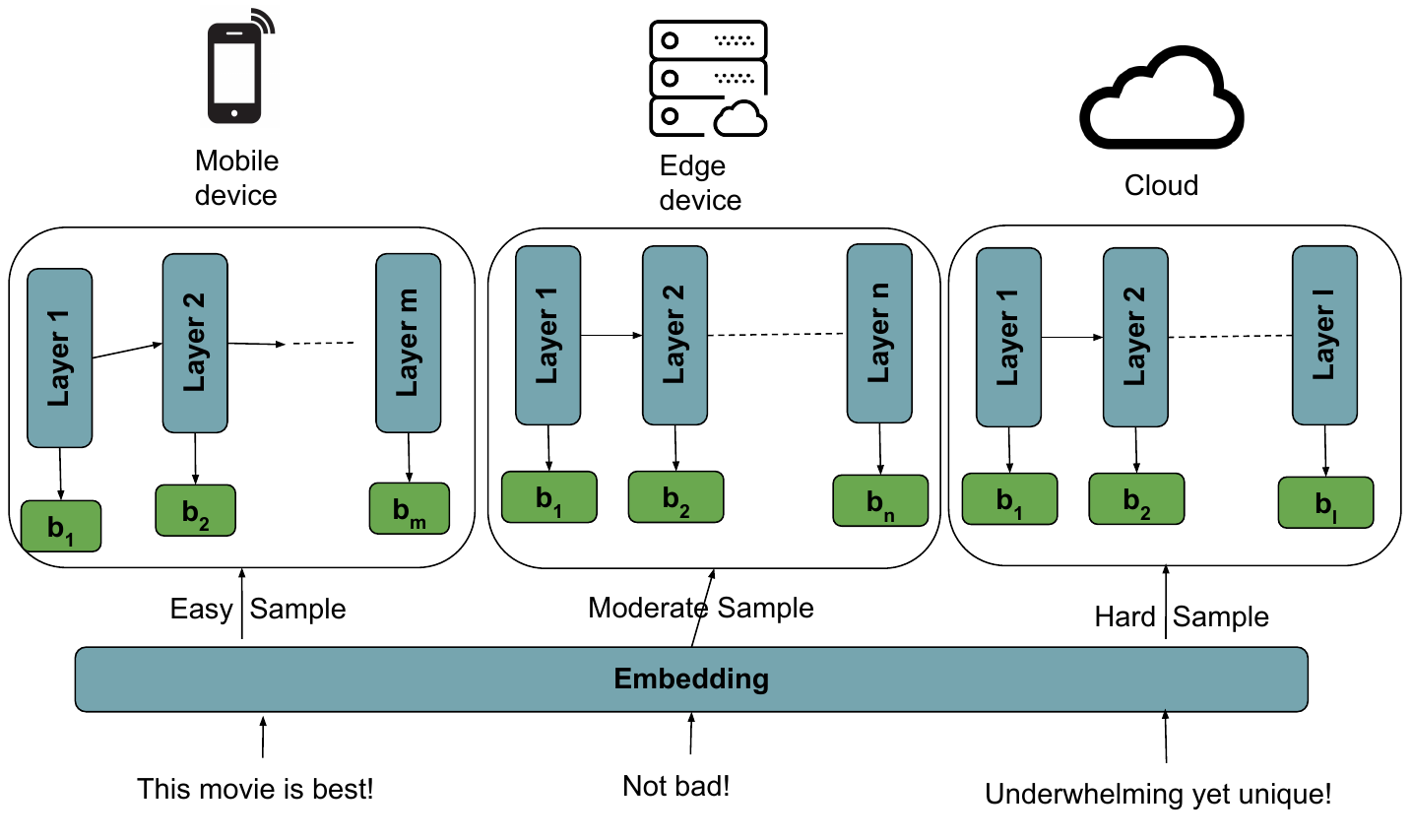}
    \caption{In this figure, three types of reviews are input to the mobile device. It passes through the embedding layer on the mobile device where it decides about the complexity of the sample. The DNN is divided into three parts: 1) First $m$ layers are deployed on the mobile device and easy samples are then inferred on the mobile device. 2) First $n$ layers are deployed on the edge device and the sample that is more complex that it can not be inferred on the mobile is inferred at the edge. 3) Finally, fully-fledged DNN is deployed on the cloud and the sample is offloaded only if it falls in the hardest pool of samples i.e. both mobile and edge cannot gain sufficient confidence to infer the sample.}
    \label{fig:main}
\end{figure*}

Our approach optimizes resource usage across mobile devices, edge devices, and the cloud by distributed inference using an early exit DNN. Three variants of the DNN are deployed: a few initial layers on the mobile device, a higher number of layers on the edge device, and a full-fledged DNN deployed on the cloud. The number of layers in every device is decided based on the available resources on the mobile and edge devices and is further discussed in Section \ref{sec:layer}. Since EE models are equipped with exit classifiers that can provide predictions on the input sample anytime, each device can independently classify incoming samples. In Figure \ref{fig:main}, we show the inference process of our method where the easier samples are processed at the mobile device, moderate are offloaded to the edge, and only the hard ones are offloaded to the cloud. 

To assess sample complexity during inference, first, we create a pool of easy, moderate and hard samples utilizing the exit points of the DNN. The pool is created during training where if a sample exits at the layer before the final layer on a mobile device, it is considered an easy sample. If a sample exits before the final layer on the edge device but after the final layer on mobile, it is classified as a moderate sample. Samples that are inferred after the final layer on the edge are considered hard samples, requiring more layers. This method gives us easy, moderate, and hard pools during training. This method effectively defines the complexity of incoming samples. During inference, we utilize the pools created during training to decide the complexity of the incoming sample. Specifically, our method analyses that the incoming sample resembles which group closely to decide its complexity on the fly using the word embeddings on the mobile device. 

In our method named \our{}: Distributed Inference on Mobile, Edge and Cloud: An Early Exit Approach, due to distributed inference, each sample is provided with an appropriate amount of computational resources as per its complexity. The major advantage of our method is that it decides on-the-fly about the computational requirements of an incoming sample without requiring it to first pass through the mobile device. This reduces the burden on individual devices and also solves the issue of inference of large models on mobile devices. Also, our method better models the accuracy-efficiency trade-off i.e., the efficiency is significantly improved in our method while maintaining the accuracy similar to that of the final layer. 
This approach effectively balances processing and communication costs, ensuring efficient and accurate processing of samples by dynamically determining whether to process them locally, at the mobile or edge, or offload them to the cloud based on sample complexity. 

For our backbone model, we adopt the well-used BERT-base/large \cite{devlin2018bert} backbone. This choice becomes an ideal test-bed for our method, given its efficiency and competitive accuracy compared to the state-of-the-art models. We conduct experiments on multiple NLP tasks to showcase the effectiveness of our method as detailed in Section \ref{sec: experiments}. Our experimental results on sentiment classification, entailment classification and natural language inference tasks demonstrate that \our{} is robust to different cost structures which means it can incorporate devices with varying processing power and communication methods such as 3G, 4G, 5G and Wi-Fi. Specifically, our method achieves a significant reduction in cost ($>43\%$) with only a minimal drop in accuracy ($<0.3\%$) when compared to the scenario where all the samples exit from the final layer.

Our key contributions are as follows:

\begin{itemize}
    \item We utilize early exits for distributed inference to enable early inferences for easy samples on mobile devices, moderate samples are inferred at the edge device and only hard samples are offloaded to the cloud.

    \item Our method is robust to various cost changes and does not lose accuracy when the mobile devices, edge devices or the communication network is changed.

    \item The minimal loss in accuracy is attributed to the fact that in our method, the optimality of the backbone is not lost i.e., we do not reduce any parameter from the backbone.

    \item We experimentally validate that our method minimizes performance degradation (sometimes even improves) while significantly reducing the costs as compared to the previous baselines.
\end{itemize}

%% file: chapters/related_works.tex
\section{Related Works}
In this section, we discuss the previous works on split computing and early exits to use DNNs on mobile devices.

\textbf{Cloud offloading} Neurosurgeon, as introduced the \cite{kang2017neurosurgeon}, explores the strategies for optimizing the splitting of DNNs based on cost considerations associated with selecting a specific splitting layer. In a similar vein, BottleNet \cite{eshratifar2019bottlenet} incorporates a bottleneck mechanism within split computing.  This approach entails deploying a segment of the DNN on a mobile device to encode the input sample into a more compact representation before transitioning to the cloud for further processing. On the same setup, multiple training strategies have been proposed for training the encoder situated on the edge device such as BottleNet++ \cite{shao2020bottlenet++} employs cross-entropy-based training approaches in the context of split computing, Matsubara \cite{matsubara2019distilled} performs knowledge distillation-based training
, CDE \cite{sbai2021cut} and Yao \cite{yao2020deep} perform reconstruction-based training 
and Matsubara \cite{matsubara2021neural} perform head-network distillation training method to effectively encode the input to offload efficiently.

\textbf{Early Exit DNNs} Early Exit DNNs have found applications across diverse tasks. In the context of image classification, BranchyNet \cite{teerapittayanon2016branchynet} and several preceding studies utilize classification entropy metrics at different intermediate layers to determine whether early inference can be made with sufficient confidence. Approaches like SPINN \cite{laskaridis2020spinn} and SEE \cite{wang2019see} incorporate early exit DNN architectures, primarily aimed at handling service disruptions in real-time inference scenarios.

Besides early exiting, works like FlexDNN \cite{fang2020flexdnn} and Edgent \cite{li2019edge} focus mainly on on the appropriate DNN depth. Other works such as DynExit \cite{wang2019dynexit}, focus on deploying the early-exit DNN in hardware. It trains and deploys the DNN on a Field Programmable Gate Array (FPGA) hardware.

In NLP domain, DeeBERT \cite{xin2020deebert}, ElasticBERT \cite{liu2021elasticbert}, CeeBERT \cite{bajpai2024ceebert}, LeeBERT \cite{zhu2021leebert} and PABEE \cite{zhou2020bert} have applied the early exit DNNs specifically for the BERT backbone. DeeBERT uses separate training to train the Early exit DNN while ElasticBERT uses the joint training strategy. CeeBERT \cite{bajpai2024ceebert} optimizes the threshold choice using multi-armed bandits. PABEE proposes a patience-based exiting criteria while LeeBERT additionally uses knowledge distillation during training. DeeCAP \cite{fei2022deecap} and MuE \cite{yang2020resolution} extend early exit ideas to image captioning models.

\textbf{DNNs on mobile device:} AdaEE \cite{pacheco2021distorted} employs a combination of Early Exit DNNs and DNN partitioning to facilitate offloading data from mobile devices to the cloud using early exit DNNs. LEE (Learning Early Exit) \cite{ju2021learning}, DEE (Dynamic Early Exit) \cite{ju2021dynamic} and UEE-UCB \cite{hanawal2022unsupervised} leverage the multi-armed bandit (MAB) framework to determine optimal exit points, while I-SplitEE \cite{bajpai2023splitee, bajpai2024splitee} also utilizes the MAB \cite{ML02_UCB1_Auer} setup to get the optimal splitting points under domain change scenarios in edge-cloud co-inference setup. LEE and DEE are specifically designed for efficient edge device inference, particularly in cases of service disruptions, employing utility functions that require ground-truth labels.

Our approach distinguishes itself from prior methods in several key aspects. 1) Our method does not require processing from one device and then offloading instead it decides on the fly, which device could fulfil sample requirements. 2) Only hard samples are offloaded to the cloud and easy ones are locally inferred at mobile and edge lowering the offloading cost.
3) We leverage the confidence in prediction to dynamically determine the level of computational resources needed for a sample.

%% file: chapters/problem_setup.tex
\section{Problem setup}
We start with a Pre-trained Language model such as BERT and attach exit classifiers after all the layers of the backbone. In the following, we discuss the setup in detail.

\subsection{Training exit classifiers}
Let $\mathcal{D}$  represent the distribution of the dataset with a label class $\mathcal{C}$ used for backbone fine-tuning. Let us assume that there are $l$ layers in the backbone. For any input sample, $(x,y)\sim \mathcal{D}$ and the $i$th exit, the loss can be computed as:

\begin{equation}
    \mathcal{L}_i(\theta) = \mathcal{L}_{CE}(f_i(x, \theta), y) 
\end{equation}

Here, $f_i(x, \theta)$ is the output of the classifier at the $i$th layer, where $\theta$ denotes the set of learnable parameters, and $\mathcal{L}_{CE}$ is the cross-entropy loss. We learn all the classifiers simultaneously hence the overall loss function could be written as $\mathcal{L} = \sum_{i=1}^l\mathcal{L}_i$. This loss simultaneously optimizes all the exits. Also, let $\hat{P}_i(c)$ denote the estimated probability class $c\in \mathcal{C}$ and $C_i$ denote the confidence in the estimate at the $i$th layer i.e., $C_i:=\max_{c\in\mathcal{C}}\hat{P}_i(c)$.
Subsequently, the model is ready for inference.

\subsection{Preparation of dataset pool}
After training, we divide the dataset into three pools based on their complexity i.e. the easy pool, the moderate pool and the hard pool. To create the pools of the datasets, we use the training and validation dataset. The complete procedure is given in Algorithm \ref{alg:pool}. 

\begin{algorithm}
\caption{Pool creation}\label{alg:pool}
\begin{algorithmic}[1]
\State {\textbf{Input:} $x$ and threshold $\alpha$}
\State \textbf{Initialize} $P_e = [ ], P_m [ ], P_h = [ ]$
\State Process the sample till layer $m$
\If{$C_i\geq\alpha$ for any $i\in \{1,2,\ldots, m\}$}
\State $P_e.append(embed(x))$
\ElsIf{$C_m<\alpha$ and $C_i\geq \alpha$ for $i \in \{m+1, \ldots n\}$}
\State Pass the sample till $n$th layer.
\State $P_m.append(embed(x))$
\Else
\State Pass the sample till the final layer.
\State $P_h.append(embed(x))$
\EndIf
\State \textbf{Return: $P_e, P_m, P_h$}

\end{algorithmic}
\end{algorithm}

It initializes three empty lists $P_e$ easy pool, $P_m$ moderate pool and $P_h$ the hard pool. As a sample arrives, it is passed through the backbone, and if the sample exits before layer $m$ i.e., $C_i\geq \alpha$ for $i\in [m]$, where the set $[m] = \{1, 2, \ldots, m\}$ we name it as an easy sample and add it to the easy pool of samples, if the sample exits at layer $n$ and not $m$ i.e., $C_m<\alpha$ and $C_i\geq\alpha$, for any layer $i$ after layer $m$ and before layer $n$ then the sample is added to the moderate pool of samples. Finally, if the sample does not exit before layer $n$, then the sample is processed till the final layer and is inferred. The sample is then added to the pool of hard samples. Finally, the algorithm returns the pool of easy, moderate and hard samples.

\subsection{Layer distribution}\label{sec:layer}
Let us assume that the mobile device contains DNN's first $m$ layers and the edge has DNN's first $n$ layers where $1\leq m\leq n\leq l$ where the cases $m=1$ is when there is no mobile device, $m=n$ means there is either no mobile or edge device. If $n = l$, it means there is no edge device. 
We discuss the impact of the values of $m$ and $n$. These values are important as they model the overall cost and are user-defined. These are used to decide the quantity of workload on different devices, i.e., mobile, edge or cloud. A higher value of $m$ means more layers are deployed on the mobile device and the processing cost e.g. battery depletion will be high, however since more layers are in the mobile device there will be a lower chance of a sample being offloaded reducing the latency cost. If the value of $n$ is high, then there will be fewer samples being offloaded to the cloud reducing the latency costs, however, it will increase load on the edge device. If both $m$ and $n$ are kept small then since less number of layers will inferred earlier, more samples will be offloaded to the cloud increasing the offloading cost and the charges of the cloud platform.

\subsection{Choice of threshold $\alpha$}
The threshold $\alpha$ used to decide the early exiting not only models the accuracy-efficiency trade-off but also impacts the cost. The cost is affected as this threshold is used to divide the dataset into three different pools. These pools are very important as they model the assignment of a sample to different devices. Hence it is very crucial to set the threshold properly. We first define the different types of costs that we consider, 1) Processing cost is the cost to process the sample through one layer of the DNN in the mobile and edge denoted as $\lambda_m$ and $\lambda_e$ respectively. 2) Offloading cost from mobile to edge and mobile to cloud denoted as $o_e$ and $o_c$ respectively. We also assume that there is a constant cost $\gamma$ charged by the cloud platform for each sample. To choose the threshold $\alpha$, we define a reward function as:

\begin{equation}\label{eq:reward}
    r(\alpha) = \left\{
        \begin{array}{ll}
            C_i-\lambda_m i & \textit{if} \quad C_{i} \geq \alpha \text{ and } i\leq m\\
             C_i- \lambda_e i-o_e & \textit{if} \quad C_{i} \geq \alpha \text{ and } m<i\leq n\\
              C_l- o_c - \gamma& \textit{ otherwise }
        \end{array}
    \right.
\end{equation}

The reward function could be interpreted as, if the sample exits at mobile device then the reward will be confidence gained subtracted by the cost of processing the sample till the $i$th layer on the mobile device. Similarly, for the edge device, the reward will be the same with an additional cost of offloading. Finally, if the sample is offloaded to the cloud, the reward will be the inference at the final layer subtracted by the cost of cloud platform and offloading cost. The expected reward function could be written as:

\begin{multline}
\mathbb{E}[r(\alpha)] = \mathbb{E}[C_i-\lambda_m i|\text{mob. inference}] P[\text{mob. inference}]\\ + \mathbb{E}[C_i-\lambda_e-o_e|\text{edge inference}] P[\text{edge inference}]\\
\mathbb{E}[C_i-\gamma-o_c|\text{cloud inference}] P[\text{cloud inference}]
\end{multline}

Now the objective is to maximize $\mathbb{E}[r(\alpha)]$ and could be expressed as $\max_{\alpha\in S}\mathbb{E}[r(\alpha)]$ where the set $S$ is the possible choices for the $\alpha$ values. Note that $P[\text{mob. inference}]$, $P[\text{edge inference}]$ and $P[\text{cloud inference}]$  is the probability that the sample will be inferred at mobile, edge and cloud respectively and depend on the value of $\alpha$.


\subsection{Post-Deployment Inference}
\textbf{Fixed:} After storing the values $P_e$, $P_m$ and $P_h$ consisting of embeddings of easy, moderate and hard samples respectively. We calculate the average of these values and name it as $P_e^a$, $P_m^a$ and $P_h^a$ respectively. The sample can be classified as easy, moderate or hard using the average values as in K-means clustering algorithm, as a sample arrives, the distance of the incoming sample is calculated from $P_e^a$, $P_m^a$ and $P_h^a$ and classifies the sample as easy, moderate or hard based on the minimum distance of the sample from the mean values of different pools. After this the easy samples are inferred locally at the mobile device incurring only processing cost, moderate samples are offloaded directly to the edge without any computation on mobile incurring small offloading cost and processing cost and the hard samples are directly offloaded to the cloud with higher offloading cost as well as cost charged by the cloud platform.

\textbf{Adaptive:} In fixed inference, the pools are created using the validation dataset, however during test time there might be a shift in the dataset distribution. For such cases, we dynamically update the pool averages such that the distribution shift can be properly captured. In this setup, as the sample arrives, it is classified as easy, moderate or hard in a similar way but this time the average is recalculated based on the incoming sample's complexity. For instance, if a sample is easy, then the value $P_e^a$ is recalculated. In this manner, the shift is captured and the trade-off of accuracy-cost is not affected. 

\begin{algorithm}
\caption{Dynamic Inference}\label{alg:inference}
\begin{algorithmic}[1]
\State {\textbf{Input:} Test sample $x$, $P_e^a, P_m^a and P_h^a$}
\State $n_e, n_m, n_h =$ number of samples in validation split. 
\State $x_e \gets embed(x)$
\State Calculate distance from all pool averages $d(x_e, \_)$
\State $d_{min}(x) \gets \min\{d(x_e, P_e^a), d(x_e, P_m^a), d(x_e, P_h^a)\}$
\If{$d_{min} = d(x_e, P_e^a)$}

\State Infer the sample locally on mobile.
\State $P_e^a\gets\frac{n_e.P_e^a+x_e}{n_e+1}$, $n_e\gets n_e+1$
\ElsIf{$d_{min} = d(x_e, P_m^a)$}
\State Offload the sample to the edge and process.
\State $P_m^a\gets\frac{n_m.P_m^a+x_e}{n_m+1}$, $n_m\gets n_m+1$
\Else
\State Offload the sample to the cloud.
\State $P_h^a\gets\frac{n_h.P_h^a+x_e}{n_h+1}$, $n_h\gets n_h+1$
\EndIf

\end{algorithmic}
\end{algorithm}

In Algorithm \ref{alg:inference}, we only show the adaptive version of our method. To obtain the fixed version from given algorithm, lines 8, 11, 14 need not be executed. However, we prefer the adaptive inference as it performs better as shown in the experiments and also comes with almost negligible computational cost.

%% file: chapters/experiments.tex
\section{Experiments}\label{sec: experiments}

\begin{table*}[]
\centering
\begin{tabular}{ccccccccccccc}
\hline
Model/Data & \multicolumn{2}{c}{SST-2}    & \multicolumn{2}{c}{CoLA}   & \multicolumn{2}{c}{MNLI}     & \multicolumn{2}{c}{MRPC}     & \multicolumn{2}{c}{QNLI}     & \multicolumn{2}{c}{QQP}    \\ \hline
           & Acc           & Cost         & Acc         & Cost         & Acc           & Cost         & Acc           & Cost         & Acc           & Cost         & Acc         & Cost         \\ \hline
BERT       & 93.5          & 1.00            & 58.3        & 1.00            & 84.5          & 1.00            & 89.2          & 1.00            & 92.5          & 1.00            & 72.1        & 1.00            \\
Random     & 89.5          & -27\%          & 55.7        & -31\%          & 79.9          & -46\%          & 86.5          & -39\%          & 89.6          & -49\%          & 69.4        & -32\%          \\
Early-Exit & 91.0            & -23\%          & 56.5        & -25\%          & 82.1          & -31\%          & 87.6          & -42\%          & 90.2          & -36\%          & 70.0          & -28\%          \\
AdaEE      & 92.1          & -36\%          & 56.9        & -40\%          & 82.8          & -42\%          & 88.1          & -51\%          & 91.4          & -41\%          & 70.8        & -30\%          \\
I-SplitEE  & 92.4          & -45\%          & 57.3        & -39\%          & 83.6          & -48\%          & 88.5          & -58\%          & 91.9          & -57\%          & 71.3        & -39\%          \\ \hline
Ours-F     & 93.3          & -43\%          & 57.8        & -42\%          & 83.9          & -53\%          & 88.9          & \textbf{-62}\% & 82.1          & -58\%          & 71.8        & -44\%          \\
Ours-D     & \textbf{93.6} & \textbf{-47}\% & \textbf{58.1} & \textbf{-43}\% & \textbf{84.3} & \textbf{-57}\% & \textbf{89.2} & -61\%          & \textbf{92.2} & \textbf{-63}\% & \textbf{72.0} & \textbf{-47}\% \\ \hline
\end{tabular}
\caption{Main results of the paper on the BERT backbone. The baseline cost is considered as the original BERT model deployed on the cloud.}
\label{tab: results}
\end{table*}

In this section, we provide all the experimental details of the paper and experimentally validate our method. 

\subsection{Dataset}
We used the GLUE \cite{wang2019glue} datasets for the evaluation of our method. We evaluate our method on three types of tasks viz. sentiment classification, entailment classification and natural language inference. The datasets used are: 

1) MRPC: Microsoft Research Paraphrase Corpus is a semantic equivalence classification dataset containing sentence pairs extracted from online news sources. 2) QQP: Quora Question Pairs is also a semantic equivalence classification dataset but the sentence pairs are extracted from the community question-answering website Quora. 3) SST-2: Stanford-Sentiment Treebank is a sentiment classification dataset. 4) CoLA: Corpus of Linguistic Acceptability with a task of linguistic acceptability of a sentence.
5) QNLI: Question-answering natural language inference is a dataset with a labelling task indicating whether the answer logically entails the question's premise. 6) MNLI: Multi-Genre Natural Language Inference also contains sentence pairs as premise and hypothesis, the task is to classify them as entailment, contradiction or neutral.

\subsection{Baselines}
We compare the model against various baselines that are detailed below:

\textbf{1) BERT model:} In this baseline, we report the results of the original BERT backbone. We assume that the BERT model is deployed on the mobile device and only processing cost is incurred. This is the main baseline for us.

\textbf{2) Random: } In this baseline, the incoming sample is randomly assigned to one of the given devices i.e. mobile, edge or the cloud. This is created to show that our assignment based on the pooling of samples makes a significant difference.

\textbf{3) Early-Exit:} is the baseline where we assume that the model is deployed completely on the mobile device. This baseline shows that splitting the model also helps due to the presence of hard samples.

\textbf{4) AdaEE:} This method is an adaptive method that uses multi-armed bandits to learn the optimal threshold to decide offloading in an edge-cloud co-inference setup. 

\textbf{5) I-SplitEE:} This method learns the optimal splitting layer based on the accuracy-cost trade-off in an online setup. The method uses multi-armed bandits to learn the optimal layer in an edge-cloud co-inference setup where the test dataset contains distortions. 

\textbf{6) Ours-F:} is our method that uses a fixed pool average and does not update it during inference.

\textbf{7) Ours-D:} is our method that dynamically updates the pool averages and covers any domain shift occurring during inference.

We use the same hyperparameters for all the baselines as given in their respective codebases. The cost for all the baselines is calculated using our cost structure which is very similar to most of the previous methods.

Following this, we detail the training and inference procedure. There are three key phases in our experimental setup.

\subsection{Training the backbone}
To evaluate our method, we use the widely accepted BERT-base/large model. We add a linear output layer after each intermediate layer of these models. We split the dataset into three parts: 80\% for training, 10\% for validation and 10\% for test. We closely follow the training procedure as described in the paper \cite{bajpai2024ceebert}. We train the backbone using the train split. We run the model for $5$ epochs. We also perform a grid search over a batch size of $\{8,16,32\}$ and learning rates of \{1e-5, 2e-5, 3e-5, 4e-5, 5e-5\} with Adam \cite{kingma2014adam} optimizer. We apply an early stopping mechanism and select the model with the best performance on the validation set.

\subsection{Pool creation and cost}
We create the pool using the validation split of the dataset. The values of $m$ and $n$ are chosen using the cost structure and we choose $m = 3, 6$ and $n = 6, 12$ for BERT-base and large respectively. The set $S$ of thresholds is chosen as ten equidistant values from $1/\mathcal{C}$ to $1.0$ where $\mathcal{C}$ denotes the number of classes. The reason for not choosing any value below $1/\mathcal{C}$ as any threshold below this value is extraneous due to the definition of the confidence values. The value $\alpha$ is chosen by solving the Equation \ref{eq:reward}.

Recall, that we have denoted the processing cost for mobile device as $\lambda_1$ and processing cost for edge device as $\lambda_2$, $o_1$ as the offloading cost for mobile to edge and the offloading cost for mobile to cloud as $o_2$. We also assume the cost charged by the cloud platform as $\mu$.

We convert all the costs in terms of the smallest unit. As we have considered the smallest cost as the processing cost of the edge device, we assume $\lambda_2 = \lambda$, $\lambda_1 = (3/2)\lambda$, $o_1 = (5/2)\lambda$ and finally $o_2 = 3\lambda$ to show the results but in the ablation studies, we experiment by varying these costs (see section \ref{sec:cost_vars}). We have fixed these values, however, we plot the accuracy cost curves when the cost values are changed. The choice of the cost values is user-specific and processing cost could be chosen as the mobile and edge device computational power and offloading costs depend on the communication networks. In Figure \ref{fig:clustering}, we show that how the pool looks like after creation. In this, we show the t-SNE (t-distributed Stochastic Neighbor Embedding) plots of the word embeddings that is used to visualize higher dimensional data.

\subsection{Inference}
During inference, we use a batch size of $1$ as data arrives sequentially. As a sample arrives, the word embedding of a sample is calculated on the mobile device. Then the distance of the word embedding of the sample is calculated against pool averages and the sample is assigned to the closest pool average. If the closest pool is the easy pool, then the sample is inferred on the mobile device. If the closest is the moderate pool, then the sample is offloaded to the edge device. Else, the sample is offloaded to the cloud. 

All the experiments were conducted on NVIDIA RTX 2070 GPU with an average runtime of $\sim$ 3 hours and a maximum runtime of $\sim$ 10 hours for the MNLI dataset.

%% file: chapters/results.tex
\section{Results}
In Table \ref{tab: results} and \ref{tab:res_large}, we show the main results of our paper, our method outperforms all the existing baselines both in terms of cost as well as accuracy for both BERT-base and large models. The reduction in call is larger for the BERT-large model which is intuitive as the large variant is more overparameterized.
\begin{table}[ht]
\centering
\begin{tabular}{ccccccc}
\hline
Model/Data & \multicolumn{2}{c}{CoLA}     & \multicolumn{2}{c}{MRPC}     & \multicolumn{2}{c}{QNLI}     \\ \hline
           & Acc           & Cost         & Acc           & Cost         & Acc           & Cost         \\ \hline
BERT       & 59.5          & 1.00         & 90.1          & 1.00         & 93.1          & 1.00         \\
Random     & 55.9          & -45\%          & 85.2          & -59\%          & 90.4          & -51\%          \\
Early-Exit & 56.2          & -32\%          & 88.2          & -48\%          & 91.2          & -55\%          \\
AdaEE      & 58.1          & -54\%          & 89.2          & -57\%          & 91.7          & -62\%          \\
I-SplitEE  & 58.4          & -57\%          & 88.9          & -58\%          & 92.4          & -68\%          \\ \hline
Ours-F     & 58.9          & -61\%          & 89.7          & -63\%          & 92.6          & -72\%          \\
Ours-D     & \textbf{59.2} & \textbf{-65\%} & \textbf{90.2} & \textbf{-67\%} & \textbf{92.8} & \textbf{-75\%} \\ \hline
\end{tabular}
\caption{Results on the BERT-large variant}
\label{tab:res_large}
\end{table}
\begin{figure*}
    \centering
    \begin{subfigure}{0.32\textwidth}
        \includegraphics[width=\textwidth]{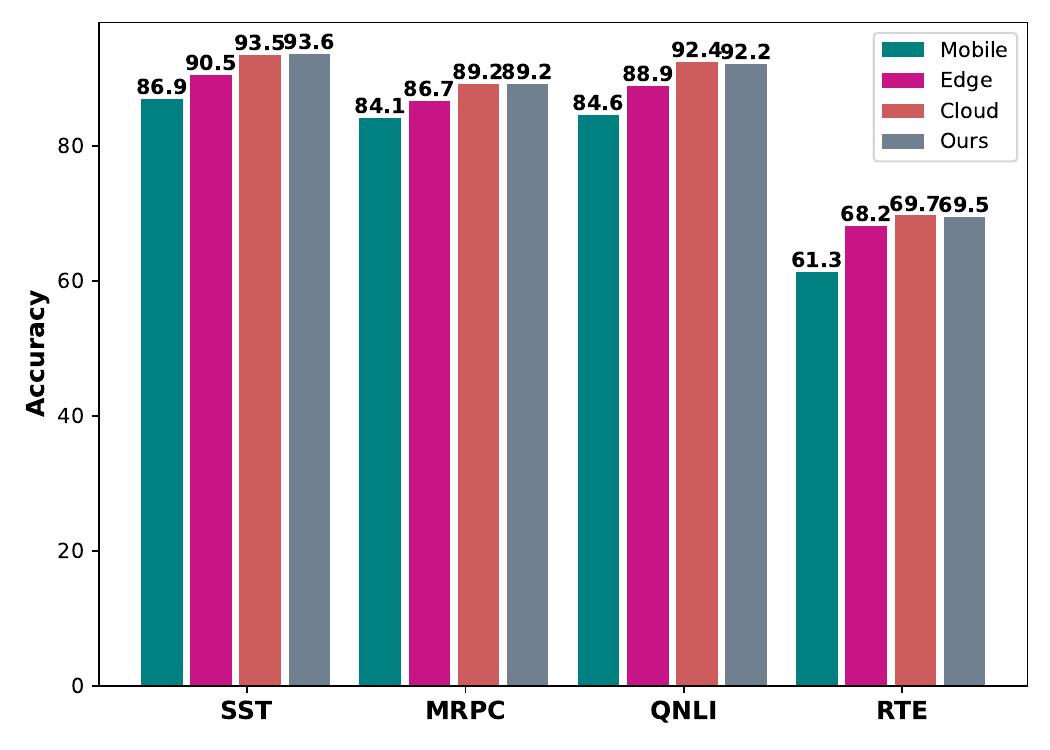}
        \caption{Accuracy of Mobile, Edge and Cloud}
        \label{fig:accuracy}
    \end{subfigure}
    \begin{subfigure}{0.32\textwidth}
        \includegraphics[width=\textwidth]{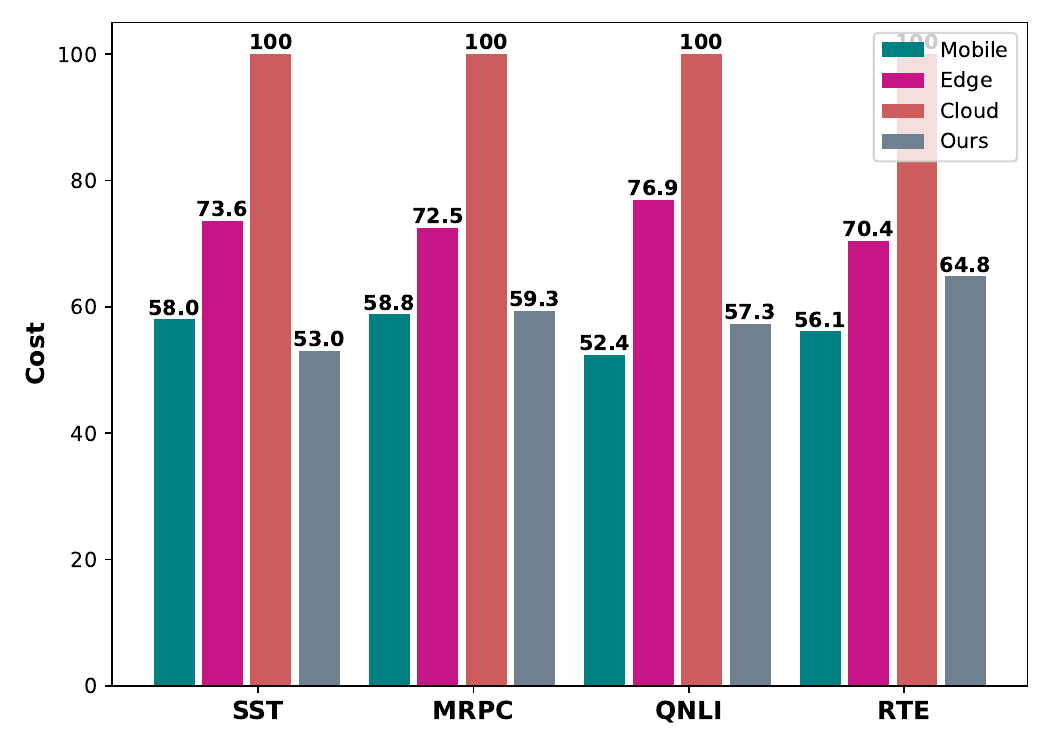}
        \caption{Cost of Mobile, Edge and Cloud}
        \label{fig:cost}
    \end{subfigure}
    \begin{subfigure}{0.32\textwidth}
        \includegraphics[width=\textwidth]{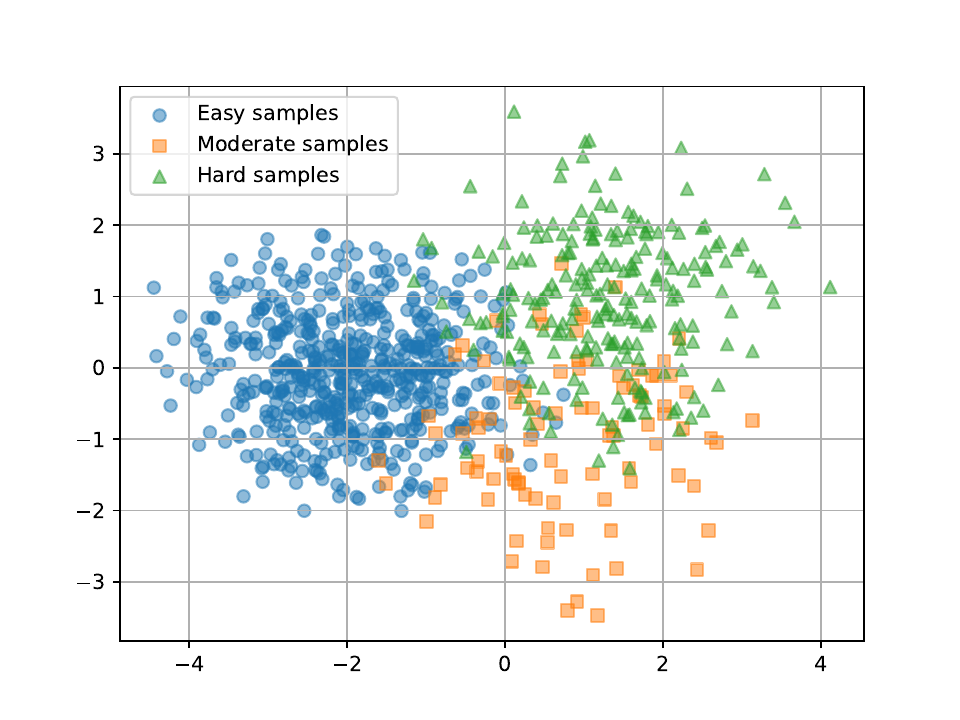}
        \caption{The t-SNE visualization of embeddings.}
        \label{fig:clustering}
    \end{subfigure}
    \caption{The figure shows the accuracy and cost of the individual devices i.e., mobile, edge and cloud. Figure on right: The t-SNE visualization of the word embeddings of the easy, moderate and hard pool created for the SST-2 dataset}
    \label{fig:global}
\end{figure*}

\begin{figure*}
    \centering
    \includegraphics[scale = 0.35]{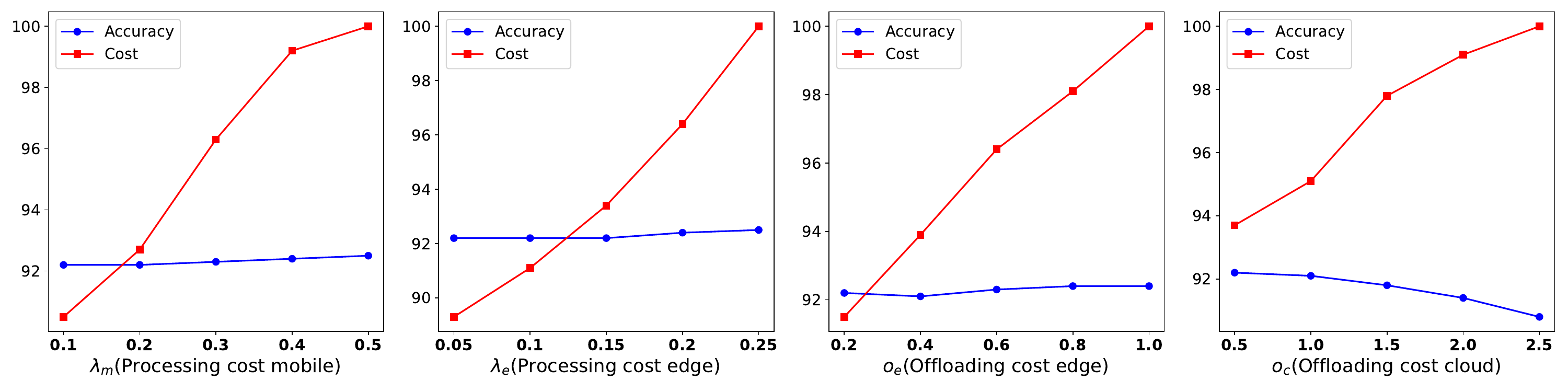}
    \caption{The changes in accuracy and percentage change in cost values when one of the cost is varied while keeping others at a constant value.}
    \label{fig:cost_variations}
\end{figure*}
The BERT model has a higher cost as all the samples are required to pass through the final layer and there are no exits attached. Due to this the accuracy of this model is comparable to our method. In the random assignment of samples to any of the devices, the loss in accuracy is due to the fact that sometimes even the hard samples are assigned to the mobile device, while the increase in cost is due to the fact that easy samples are sometimes assigned to the cloud. The vanilla early exit method gets a lower accuracy as the threshold chosen for exiting is not chosen based on any optimization algorithm but only through a validation set. This signifies the importance of the choice of threshold using the reward function \ref{eq:reward}. AdaEE also has lower accuracy as the method mostly works better under domain change scenarios, but in our case the domain shift is minor, it simply reduces to an early exit model with dynamic learning of threshold. Due to this dynamic learning of threshold, it outperforms vanilla early exiting. Finally, the I-SplitEE model also has lower accuracy again due to the case that, it works better in domain shift scenarios. In terms of cost, these models are higher as they require all the samples to be processed on the mobile device before offloading.

Our method outperforms all the baselines, the higher accuracy comes from the appropriate assignment of the sample to various devices and a smaller cost as compared to other methods since all the complexity of the sample is decided based on the word embedding that does not require much processing on mobile reducing the processing cost to a larger extent while in other methods, this cost is really high as it is not removed for any sample.

Also, note that for some datasets our method even outperforms the vanilla BERT inference, this is because of the overthinking issue during inference. This issue occurs when an easy sample is passed through the complete backbone leading to the extraction of irrelevant features which in turn results in a wrong prediction as pointed out in \cite{zhou2020bert}.



\section{Ablation Study and Discussion}
In this section, we perform ablation studies and also discuss the choice of layers using the computational powers of mobile and edge and offloading costs.
\subsection{Individual device inference}
We stated that our method uses a distributed inference method between mobile, edge and cloud. In Figure \ref{fig:global}, we show the effect on cost and accuracy when all the samples are inferred on one of the given devices. It means that instead of distributing the inference, performing the inference on a single device. In figure \ref{fig:accuracy}, we plot the accuracies of the individual devices and our model. Since the cloud contains the full-fledged DNN, it has the highest accuracy; however, our method sometimes outperforms the full-fledged DNN due to the overthinking issue in DNNs. In terms of cost, we know that the highest cost will be of the cloud. Hence, the cost in figure \ref{fig:cost} is given in terms of the percentage of cost saved as compared to the cloud. Our method has a slightly higher cost than only mobile setup as it involves offloading of samples. 
Also, note that for easier tasks such as sentiment classification, most of the samples are inferred on the mobile device while for harder tasks such as entailment classification, more samples are offloaded hence a higher cost.

\subsection{Cost Variations}\label{sec:cost_vars}
In Figure \ref{fig:cost_variations}, we show the variation in accuracy and cost if we alter one of the given costs. In the left figure in Figure \ref{fig:cost_variations}, we alter $\lambda_m$ i.e., the processing cost of mobile device, while keeping other costs constant. The accuracy is not affected in this case as we are increasing the processing cost that forces more samples to offload to the edge and cloud having more layers hence accuracy slightly improves. Similarly, if the processing cost $\lambda_e$ is increased, accuracy again slightly improves. As we increase the offloading cost of the cloud $o_c$, we observe that there is a drop in accuracy. This is expected as higher offloading costs for the cloud will set lower thresholds such that most of the samples are inferred locally or at the edge and do not offload to the cloud which in turn lowers the accuracy. Note that in this setup, other costs are kept constant. Also, our method is robust to changes in different types of costs i.e., the loss in accuracy is minimal when cost values vary.

%% file: chapters/conclusion.tex
\section{Conclusion}
We address the inference of large DNNs on mobile devices using the complexity of the input samples. We propose a method that utlizes early exits to decide the complexity of samples. It minimizes the cost of inference by assigning appropriate amount of resources required to infer the incoming sample between mobile, edge and cloud. If the task is easy and require less computation then most of the samples are inferred locally while if the task is hard, then most of the samples are offloaded maintaining accuracy. Our method is robust to changes in cost values i.e., various mobile and edge devices. Experiments on various NLP tasks show the significance of our work where the drop in accuracy is $(<0.3\%)$ while reducing the cost $(>43\%)$ as compared to final layer on the cloud.